\pdfoutput=1

\documentclass[11pt]{article}

\usepackage[final]{acl}

\usepackage{times}
\usepackage{latexsym}

\usepackage[T1]{fontenc}

\usepackage[utf8]{inputenc}

\usepackage{microtype}

\usepackage{inconsolata}

\usepackage{graphicx}
\usepackage{epsfig} 
\usepackage{booktabs}
\usepackage{array}
\usepackage{makecell}
\usepackage{tabularx}
\usepackage[utf8]{inputenc}
\usepackage{multirow}
\usepackage{float}
\usepackage{caption}
\usepackage{geometry}
\usepackage{icomma}
\usepackage{enumitem}
\usepackage{kotex}
\usepackage{amssymb}
\usepackage{tabularray}
\usepackage{amsmath}
\usepackage{tabularx}
\usepackage{stackengine}
\usepackage{courier}
\usepackage[T1]{fontenc}
\usepackage{hyperref}
\usepackage[fixed]{fontawesome5}
\usepackage{adjustbox}

\newcommand{\succprec}{\mathrel{\vcenter{\hbox{$\stackMath\stackinset{l}{}{c}{1.4ex}{\prec}{\succ}$}}}}
\newcommand{\ie}{\textit{i}.\textit{e}.}
\newcommand{\eg}{\textit{e}.\textit{g}.}

\newcommand{\kyungjae}[1]{\textcolor{blue}{#1}}
\newcommand{\Red}[1]{\textcolor{red}{#1}}

\DeclareMathOperator*{\argmax}{arg\,max}

\title{Reinforcement Learning from Reflective Feedback (RLRF): \\ Aligning and Improving LLMs via Fine-Grained Self-Reflection}

\newcommand\firstauthor{\thanks{~Equally contributed to this work.}}

\author{Kyungjae Lee\firstauthor \quad\quad Dasol Hwang$^*$ \quad\quad  Sunghyun Park$^*$ \\ \textbf{Youngsoo Jang  \quad\quad Moontae Lee}  \\  LG AI Research \\
\texttt{\{kyungjae.lee,dasol.hwang,sunghyun.park},\\\texttt{youngsoo.jang,moontae.lee\}@lgresearch.ai}
}

\begin{document}
\maketitle

\begin{abstract}

Despite the promise of RLHF in aligning LLMs with human preferences, it often leads to superficial alignment, prioritizing stylistic changes over improving downstream performance of LLMs. Underspecified preferences could obscure directions to align the models. Lacking exploration restricts identification of desirable outputs to improve the models. To overcome these challenges, we propose a novel framework: Reinforcement Learning from Reflective Feedback (RLRF), which leverages fine-grained feedback based on detailed criteria to improve the core capabilities of LLMs. RLRF employs a self-reflection mechanism to systematically explore and refine LLM responses, then fine-tuning the models via a RL algorithm along with promising responses. Our experiments across Just-Eval, Factuality, and Mathematical Reasoning demonstrate the efficacy and transformative potential of RLRF beyond superficial surface-level adjustment.
\end{abstract}
\section{Introduction}

Reinforcement Learning from Human Feedback (RLHF) has emerged as a crucial framework for aligning large language models (LLMs) with human preferences. To facilitate preference alignment, existing approaches such as InstructGPT~\cite{NEURIPS2022_b1efde53}, Sparrow~\cite{glaese2022improving}, Llama-2~\cite{touvron2023llama} commonly train a reward model with preferential human feedback. This reward model assesses the overall quality of model outputs as a scalar value. Then training LLMs with the reward signals encourages the models to generate more favorable responses better aligned with human preferences.

Despite recent successes in preference alignment, training LLMs through RLHF does not guarantee a significant improvement of LLM's capabilities, in terms of downstream performance in NLP tasks. Previous works \cite{zhou2023lima,lin2023unlocking} have raised skepticism regarding the efficacy of current alignment techniques in improving LLM's capabilities. \citet{zhou2023lima} claim that such alignment tuning might be \textit{superficial} learning, where the model primarily learns favorable styles or formats for interacting with users. \citet{lin2023unlocking} also observe that most distribution shifts between base and post-alignment LLMs tend to be predominantly in stylistic tokens. However, enhancing the capabilities of LLMs is more critical than adjusting their interaction styles or formats to better match human preferences.

To address the superficial nature of preference alignment, we first investigate why the current RLHF often leads surface-level alignment. We tackle factuality and mathematical reasoning because the stylistic adjustment rarely contributes to downstream performance. Observing preference-based reward models is notably deficient in evaluating mathematical reasoning, we hypothesize that \textit{preference-based} reward models may cause superficial alignment. As a solution, we leverage \textit{fine-grained} LLM feedback that incorporates both verbal response and numeric score adhering to detailed criteria. However, even if adopting RL fine-tuning with fine-grained feedback as a reward, improving LLM capabilities remains a significant challenge due to the combinatorial action space, the vast array of potential responses in NLP tasks~\cite{ramamurthy2023is,yehudai-etal-2022-reinforcement,zhuang2023toolchain}.

\begin{figure*}[t]
	\centering
	\includegraphics[width=162mm]{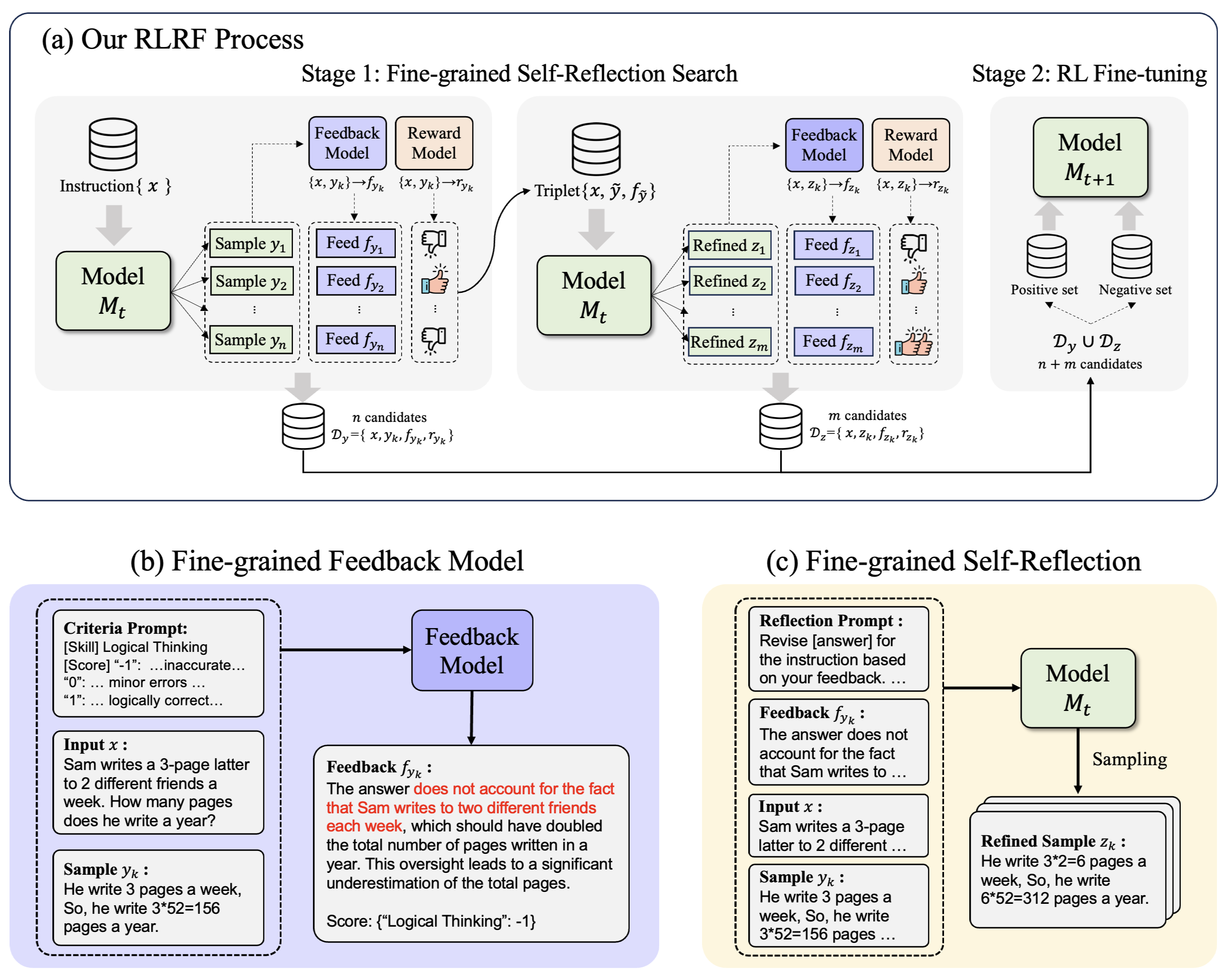}
	\caption{An overview of our proposed Reinforcement Learning from Reflective Feedback (RLRF).}
	\label{figure1}
\end{figure*}

To this end, we introduce a novel framework: \textbf{Reinforcement Learning from Reflective Feedback (RLRF)}, designed to effectively explore promising responses and improve LLM capabilities through fine-grained feedback. \textit{Self-reflection}, which empowers LLMs to evaluate and refine their responses based on feedback against previous outputs \cite{madaan2023self,ganguli2023capacity,welleck2022generating,pan2023automatically,chen2023teaching}, is the key idea that enables targeted exploration on promising responses. High-quality outputs that have been improved through self-reflection lead to advance LLM capabilities with RL fine-tuning.

Our framework consists of  the following two stages as illustrated in Figure~\ref{figure1}. Initially, the \textbf{Fine-Grained Self-Reflection} stage exploits the self-reflection ability of LLMs along with a fine-grained feedback model to search refined responses with high-quality. Then the \textbf{RL Fine-tuning} stage applies a RL algorithm to fine-tune the LLM utilizing these refined responses and their associated scores. In the experiments, we assess our approach on LLM-based evaluation benchmarks including Just-Eval~\cite{lin2023unlocking}, Factscore~\cite{min-etal-2023-factscore}, and GSM8k~\cite{cobbe2021training}. We employ the Llama-2 13B model~\cite{touvron2023llama} after fine-tuning on the customized open-source instruction data (See Table~\ref{table:dataset}). Note that the RLRF framework is flexible and scalable. Users can iterate the Fine-Grained Self-Reflection stage multiple times to attain higher-quality responses. RL Fine-tuning stage is not limited to applying only preference-based approaches, while our experiments are based on Direct Preference Optimization (DPO)~\cite{rafailov2023direct}.

\section{Preliminaries}

\subsection{Preference-based RLHF}\label{baseline}

Preference-based RLHF aims to optimize the policy (\ie, LLM) that aligns with human preferences using the pre-collected pairwise preference dataset $\mathcal{D}=\{(x^i,y^i_+,y^i_-)\}_{i=1}^{N}$, where $x^i$ is instruction, and $y^i_+$ and $y^i_-$ indicate the chosen and rejected responses, respectively. Conventional RLHF methods~\cite{NEURIPS2022_b1efde53,glaese2022improving} train a preference-based reward model (RM) on the pairwise preference dataset, then optimize the policy using the trained reward model. The reward model is trained by a binary ranking loss with respect to the pairwise dataset $\mathcal{D}$ as follows:
\begin{equation}
    \mathcal{L}_{\text{RM}}(\theta) = - \mathbb{E}_{\mathcal{D}} \big[\log(\sigma(r_\theta(x,y_+) - r_\theta(x,y_-)))\big],
    \label{eq:reward_learning}
\end{equation}
where $\sigma$ is the logistic function, $r_\theta(x, y)$ is the scalar output of reward model for instruction $x$ and response $y$ with parameters $\theta$, and $y_+$ and $y_-$ indicate the chosen and rejected responses, respectively. Then, the policy is optimized by the following KL-penalized objective that combines the learned reward and KL-divergence between the current policy and reference policy:
\begin{equation}
    \max_\phi \mathbb{E}_{x \sim \mathcal{D}, y \sim \pi_\phi(y|x)} \bigg[r_{\theta}(x,y) - \beta \log \frac{\pi_\phi(y|x)}{\pi_{\text{ref}}(y|x)}\bigg],
    \label{eq:policy_optimization}
\end{equation}
where $\pi_\phi$ is the policy (\ie, LLM) with parameters $\phi$, $\pi_{\text{ref}}$ is the reference policy (\eg, initial policy), and $\beta$ is a coefficient that balances the trade-off between the learned reward and the KL penalty. This KL-penalized objective can mitigate \textit{overoptimization}~\cite{gao2022scaling} of the learned reward model, and is commonly optimized by proximal policy optimization (PPO)~\cite{schulman2017proximal}.

One of the recent notable preference-based RLHF algorithms, direct preference optimization (DPO)~\cite{rafailov2023direct}, directly optimizes the policy from the pre-collected pairwise preference dataset $D$ without explicit training of reward model. \citet{rafailov2023direct} show that training of reward model (Eq.~\eqref{eq:reward_learning}) and policy optimization (Eq.~\eqref{eq:policy_optimization}) processes can be replaced by optimizing the following simple binary classification objective on the pairwise preference dataset $D$:
\begin{align}
    \mathcal{L}_{\text{DPO}}(\phi) = - \mathbb{E}_{\mathcal{D}} \bigg[\log\sigma\bigg(&\beta \log \frac{\pi_\phi(y_+|x)}{\pi_{\text{ref}}(y_+|x)} \nonumber \\ 
    & - \beta \log \frac{\pi_\phi(y_-|x)}{\pi_{\text{ref}}(y_-|x)} \bigg)\bigg]. 
    \label{eq:dpo_objective}
\end{align}
This single-stage policy learning of DPO enables more stable and efficient training, compared to PPO.

\subsection{Challenges in Improving the Capabilities of LLM via preference-based RLHF}
Despite recent successes of RLHF, fine-tuning LLMs with RLHF still has many challenges such as instability of training~\cite{zheng2023secrets}, sensitivity to hyperparameters~\cite{ramamurthy2023is}, and overoptimization~\cite{gao2022scaling} of the learned reward model. 
Unlike the prior works that address the conventional challenges in RLHF, we focus on the following challenges of preference-based RLHF that are relevant to improving the capabilities of LLMs:
\begin{itemize}[leftmargin=0.4cm]
    \item \textbf{Underspecified Preference Criteria}: Several works \cite{bansal2023peering,wu2023fine,krishna2023longeval,ye2023flask} show that it is challenging for human annotators to consistently evaluate the overall quality of responses due to their different criteria for multiple aspects. Thus, to achieve the improvement of specific capabilities of LLMs, the \textit{fine-grained evaluation ability} of specific aspects is essentially required.
    \item \textbf{Restricted Exploration}: 
    One of the major challenges of RL finetuning on LLMs is combinatorial action space in NLP tasks. 
    Due to this complexity, it is infeasible to find an optimal policy through the exploration based on a naive exhaustive search.
    Previous RLHF approaches commonly used temperature-based sampling for exploration, to sample diverse outputs by increasing token-level randomness. To reduce the search space in language generation, top-$k$ sampling~\cite{fan-etal-2018-hierarchical} and nucleus sampling~\cite{Holtzman2020The} could be alternatives, but these methods still have difficulty in \textit{exploring high-quality responses}.
\end{itemize}

\section{RL from Reflective Feedback (RLRF)}
In this section, we introduce Reinforcement Learning from Reflective Feedback (\textbf{RLRF}), a framework designed to produce promising responses through self-reflection, then improve the capabilities of LLMs with RL fine-tuning. Specifically, we present the \textbf{Fine-Grained Feedback Model} (Sec \ref{fine_grained_feedback_model}), which can criticize the responses and evaluate the fine-grained capabilities of LLMs in multiple aspects (\eg, logical correctness, factuality, insightfulness). Then we will describe Reinforcement Learning from Reflective Feedback (\textbf{RLRF}), which consists of the following two components that leverage the fine-grained feedback model: (1) \textbf{Fine-Grained Self-Reflection}, which exploits LLM's self-reflection capability with fine-grained feedback model to search high-quality refined responses (Sec \ref{fine_grained_self_reflection}), (2) \textbf{RL Fine-tuning}, which fine-tunes the LLM on the refined dataset with the RL algorithm (Sec \ref{RL_fine_tuning}).

\subsection{Fine-Grained Feedback Model}
\label{fine_grained_feedback_model}
To address the first challenge of \textit{underspecified criteria},
we present a fine-grained feedback model, which can evaluate the responses from LLMs on fine-grained criteria for multiple aspects.
Prior studies have shown the limitations of evaluating LLMs' responses with a single metric of preference \cite{bansal2023peering,wu2023fine,krishna2023longeval,ye2023flask}. Recently, \citet{ye2023flask} have developed a fine-grained language model evaluation method for the capabilities of LLMs using LLM as an evaluator. Inspired by this, we define the following eight evaluation aspects with three-level rating rubrics in each aspect: Factuality, Logical Correctness, Metacognition, Insightfulness, Completeness, Comprehension, Readability, Harmlessness (See Table \ref{table:skill_prompt}). In defining the rating rubrics, we focus on recognizing whether the response $y$ meets specific standards (categorized as success, moderate, or failure), whereas previous works \cite{liu2023gpteval,zheng2023judging} employed a wide range of rating scales, such as 5 or 10 points.

To achieve focused evaluation on aspects that are essential to follow each instruction, our feedback model selects the top-3 relevant aspects from the whole aspect set and then evaluates the selected aspects, similar to the approach proposed by \cite{ye2023flask}. Finally, our fine-grained feedback model $f_{p}$ with rubrics for all aspects as prompt $p$, generates the feedback $f_{p}(x,y)$ on three relevant aspects (See Table~\ref{table:prompt_feedback1}). We parse per-aspect ratings in the last sentence of $f_p(x,y)$, and use the ratings to complement the underspecified reward $r(x,y)$ (\ie, preference-based reward). 
For brevity, we will refer to the fine-grained feedback model and preference-based reward model as the feedback model and reward model in the remaining sections.

Optionally, if the task of a given instruction is known, we can evaluate on a single task-specific aspect (See Table \ref{table:prompt_feedback2}). For task-specific instructions, we align them with a single fixed aspect. For example, a mathematical reasoning task can be aligned with ``logical correctness'', while aligning a biography generation task with ``factuality''. 
In such NLP tasks, if reference knowledge or answers are available, we can boost the critique capabilities of feedback models prompting with the reference, which enables the feedback to be grounded in the reference (See Sec \ref{feedback_analysis}).
We used Wikipedia articles for a biography generation task, and human answers for a mathematical reasoning task. 

\subsection{Fine-Grained Self-Reflection}
\label{fine_grained_self_reflection}

To tackle the second challenge of \textit{restricted exploration}, we present a fine-grained self-reflection, which can effectively explore high-quality responses among the massive set of available responses. Unlike other RLHF approaches that explore diverse outputs through temperature-based sampling, we encourage effective exploration by leveraging the LLM's self-reflection ability that provides feedback and uses it to refine itself. To boost self-reflection, we employ the feedback $f_p(x,y_k)$ as such prompt which provides detailed reasons behind the model's mistakes, facilitating more effective reflection and improvement.

Fine-grained self-reflection starts by selecting a promising response $\tilde{y}$ to be refined. To select a promising response, we generate a set of $n$ candidate responses for given instruction $x$ and their evaluations as follows:
\begin{equation*}
    \mathcal{D}_y=\{(x,y_i,f_p(x,y_i),r(x,y_i))| y_i\sim\pi_\phi(x) \}_{i=1}^n,
\end{equation*}
where $\pi_\phi$ is the policy (\ie, LLM) and $y_i$ is generated response by temperature-based sampling. Then, $\tilde{y}$ is selected as the promising response with the highest preference-based reward among the candidate responses as follows:
\begin{equation*}
    \tilde{y}=\argmax_{y\in\{y_i\}_{i=1}^n} r(x,y),
\end{equation*}
where $\{y_i\}_{i=1}^n$ is a set of $n$ response candidates. 
To effectively explore high-quality responses, we generate $m$ refinement by performing self-reflection that reads the feedback $f_p(x,\tilde{y})$ and corrects the errors in $\tilde{y}$ as follows:
\begin{equation*}
    \mathcal{D}_z=\{(x,z_j,f_p(x,z_j),r(x,z_j))| z_j\sim\pi_\phi(\tilde{x})\}_{j=1}^m,
\end{equation*}
where $\tilde{x}=\{x,\tilde{y}, f_{p}(x,\tilde{y})\}$ and $z_j$ is refined response by self-reflection.
We use both generated datasets $\mathcal{D}_y$ and $\mathcal{D}_z$ in RL fine-tuning.

\begin{table*}[ht!]
\centering
\renewcommand{\arraystretch}{1.25}
\scalebox{0.85}{
\begin{tabular*}{2.25\columnwidth}{c|c|c|l}
\toprule
Type & Data Size & Data Format & ~~~~~~~~~~~~~~~~~~~~~~~~~~~~~~~~~~~~~~ Data Name  \\ 
\midrule

\multirow{2}{*}{ \makecell{SFT Seed for \\ Initial $M_0$} }      & 100K  & $x \mapsto y$  & \textbullet\ UltraChat, Airoboros, Open-Orca, Open-Platypus 
\\ \cline{2-4}
& 23K & $(x,y,f) \mapsto \tilde{y}$ & \textbullet\ Reflection Custom  \\ \midrule

\makecell{Preference-based\\Reward Model}     & 550K  & $(x,y_a) \succprec (x,y_b)$  & 
\makecell[l]{\textbullet\ Anthropic HH, OpenAI Summarize, WebGPT, \\ StackExchange, Stanford SHP, UltraFeedback}
\\ \midrule

\makecell{Task-augmented  \\Reward Model}     & \makecell{550K \\ + 23K}  & $(x,y_a) \succprec (x,y_b)$  & 
\textbullet\ Preference Data (550K) + Math (16K) + Factuality (7K)
\\ \midrule

\multirow{3}{*}{\makecell{Feedback \\Model}}   & 30K & $(x,y) \mapsto f$ & \textbullet\ Instruction-following Custom (sampled from SFT Seed) \\\cline{2-4}
& 9K & $(x,y, \textsc{GT}) \mapsto f$ & \textbullet\ Math Custom on GSM8K and MATH  \\\cline{2-4}
& 8K & $(x,y, \textsc{Ref} ) \mapsto f$ &  \textbullet\ Factuality Custom (Biography generation)
\\ \midrule

\multirow{3}{*}{RL fine-tuning} & 60K  & $(x, y_{+}, y_{-})$ & \textbullet\ ShareGPT \\\cline{2-4}
& 10K & $(x, y_{+}, y_{-})$ & \textbullet\ Math Custom on GSM8K and MATH  \\\cline{2-4}
& 10K & $(x, y_{+}, y_{-})$ &  \textbullet\ Factuality Custom (Biography generation) \\

\bottomrule
\end{tabular*}
}
\caption{Training data for our reward, feedback, and policy models. We list both the open-source dataset and custom data collected by GPT-4's API. The more details of this training data can be found in Appendix \ref{appendix_data}.
}
\label{table:dataset}
\end{table*}

\subsection{RL Fine-tuning}
\label{RL_fine_tuning}
In the last stage, we fine-tune the language model (\ie, policy $\pi_\phi$) via DPO~\cite{rafailov2023direct} which is one of the representative RL algorithms for fine-tuning LLMs. 
Since DPO directly optimizes the policy from the pairwise preference dataset, it requires positive-negative pairs in the form of comparable preference. We construct positive-negative pairs with whole datasets $\mathcal{D}=\mathcal{D}_y \cup \mathcal{D}_z$, which are generated from the fine-grained self-reflection stage. 
First, we classify the dataset into positive dataset $\mathcal{D}_+$ and negative dataset $\mathcal{D}_-$ through the feedback score. 
The responses with ratings of all aspects in feedback $f_p(x,y)$ being 1 (\ie, ``success'') are selected as the positive set, and the remaining responses that include the rating of 0 (``moderate'') or -1 (``failure'') are selected as the negative set. 
Among the examples in $\mathcal{D}_+$, we select top-$k$ responses with the highest reward as the positive examples $y_+$. For 1-to-1 pair matching, we randomly sample negative examples $y_-$ from $\mathcal{D}_-$ according to the number of the positive set, and discard examples with no positive set. 
Finally, we fine-tune the language model by optimizing the DPO objective (Eq.~\eqref{eq:dpo_objective}) by leveraging the $y_+$ and $y_-$ as pairwise datasets. We use DPO~\cite{rafailov2023direct} for the RL fine-tuning stage, but our framework is not limited to applying only preference-based approaches such as DPO.

\subsection{Iterative Training}
\label{iterative_training}

Figure~\ref{figure1} summarizes the overall process and details of our proposed framework. Our framework serves the iterative training that alternates between fine-grained self-reflection and RL fine-tuning. Since the updated policy can generate better responses and refinements during the fine-grained self-reflection process than the outputs from the previous policy, policy improvement can be continuously performed by repeating this process until the policy performance converges.

\section{Experiment}

\subsection{Experimental Setup}

\paragraph{Training Dataset}
Table \ref{table:dataset} summarizes the training data.
Starting with the base model, we fine-tune three independent models (the feedback, reward, and initial policy models) on the open-sourced dataset and our additional custom dataset extracted from OpenAI API (gpt-4-1106-preview). During RL fine-tuning, we use the following three types of datasets: (1) general instruction: ShareGPT, 
(2) Math reasoning: GSM8K \cite{cobbe2021training}, MATH \cite{hendrycks2021measuring}, (3) Factuality \cite{min2023factscore}.
We provide more details regarding these datasets in Appendix \ref{appendix_data}.

\begin{table*}[ht!]
\centering
\scalebox{0.92}{%
\begin{tabular}{l|c|cccc|c|c}
\toprule
\multicolumn{1}{c|}{\multirow{2}{*}{\textbf{Method}}}& \multicolumn{5}{c|}{\textbf{Just-Eval} (by GPT-4)} & \multicolumn{1}{c|}{\multirow{2}{*}{\textbf{~~FactScore~~}}} & \multicolumn{1}{c}{\multirow{2}{*}{\textbf{~~\makecell{Math\\Accuracy}~~}}}  \\ 
 & ~~~\textbf{Total}~~~ & Helpful & Depth & Factual & Math  & &   \\ 
\noalign{\hrule height 1pt}
\multicolumn{7}{l}{\textit{\textbf{SOTA LLMs}}} \\
\hline
~ GPT-4-0613  & 4.80 & 4.86 & 4.49 & 4.49 & 5.00 & 83.20 & 94.60$^{\dag}$  \\
~ GPT-3.5-turbo-0301 & 4.75 & 4.81 & 4.33  & 4.33 & 5.00  & 79.00 & 80.80$^{\dag}$  \\
~ Llama-2-70b-chat & 4.72 & 4.58 & 4.38 & 4.38 & 3.12 & 67.70 & 56.80$^{\ddag}$  \\
~ Llama-2-13b-chat & 4.45 & 4.41 & 4.02 & 4.24 & 2.38 & 65.30 & 43.14 \\
\hline
\multicolumn{7}{l}{\textit{\textbf{Our RLRF}}} \\
\hline
~ Initial $M_0$ &  4.60 & 4.58 & 4.17 & 4.51 & 4.00 & 70.79 & 41.77  \\
~ $M_1$ (RS) &  4.65 & 4.63 & 4.24 & 4.54 & 3.44 & 72.20 & 47.84  \\
~ $M_1$ (DPO) &  \textbf{4.66} & \textbf{4.66} & \textbf{4.27} & \textbf{4.55} & 3.88 & 78.50 & 47.92  \\
~ $M_2$ (RS $\rightarrow$ \text{DPO}) & 4.64 & 4.62 & 4.23 & \textbf{4.55} & 3.75 & 76.30 & \textbf{51.02}  \\
~ $M_2$ (DPO $\rightarrow$ \text{DPO}) & 4.63 & 4.63 & 4.24 & 4.52 & \textbf{4.06} & \textbf{79.30} & 49.66 \\
\hline
\multicolumn{7}{l}{\textit{\textbf{RLHF} Baseline}} \\
\hline
~ $M_1$ (RS, Reward-only) & 4.63 & 4.59 & 4.23 & 4.49 & 3.19 & 69.10 & 39.27 \\
~ $M_1$ (DPO, Reward-only) & 4.62 & 4.60 & 4.19 & 4.53 & 3.44 & 70.79  & 41.09 \\
\bottomrule        
\end{tabular}%
}
\caption{The main results of RLRF compared to various open and closed models. The best results among 13B-based models are \textbf{bold}-faced. The dagger ($\dag$) indicates the results on the CoT setting reported in \cite{zhao2023automatic}, while the double dagger ($\ddag$) is the result on 8-shot setting reported in \cite{touvron2023llama}.} 
\label{table:main_result}
\end{table*}

\paragraph{Base Model and Hyper-parameters}
In our experiment, we used the Llama-2-13b-chat as our base model. All experiments were conducted on 16 A100 GPUs, each with 40 GB of memory. We set the learning rate to 2e-5 (constant) for fine-tuning both the feedback and the initial policy models, and 2e-6 (cosine decay) for DPO.
In DPO fine-tuning, we set $\beta$ = 0.1, in Eq (\ref{eq:dpo_objective}). 
The results of additional values ($\beta$ = 0.01, 0.1, 0.5) can be found in Appendix \ref{hyperparameter_beta}.
In the self-reflection stage, we restricted the maximum samples of exploration to $n+m=30$, where $n$ is the size of $\mathcal{D}_y$ and $m$ is that of $\mathcal{D}_z$. 
In our experiments, we select the best values ($n=10$, $m=20$) among (10, 20), (15, 15), and (20, 10) on a subset of training set.

\paragraph{Baselines for Comparison}
In our experiment, we take state-of-the-art LLMs: GPT-4-0613, GPT-3.5-turbo-0301, and Llama-2-70b-chat.
As RLHF baselines, we used only the reward model without our feedback model, learning on pairs of positive examples with highest reward and random negative examples in $\mathcal{D}_y$.
As an alternative to DPO using both positive and negative examples, we can supervised fine-tune the model on only the positive set, which we call Rejection Sampling (RS).

\subsection{Evaluation Benchmarks}
To measure the effectiveness of our RLRF in multi-aspects, we conduct experiments on Just-Eval~\cite{lin2023unlocking} for fine-grained evaluation by GPT-4. This benchmark consists of 1,000 instructions from diverse datasets including AlpacaEval~\cite{alpaca_eval}, MT-bench~\cite{zheng2023judging}, LIMA~\cite{zhou2023lima}, HH-RLHF-redteam~\cite{ganguli2022red}, and MaliciousInstruct \cite{huang2023catastrophic}. This benchmark provides the categories of task type and topic for each example, which enables comprehensive analysis over diverse categories.
We report four metrics in Just-Eval: Total (avg. six aspects), Helpfulness, Depth, Factuality, and Mathematics (helpfulness over math problems). 
Our complete results, including those specific to aspects, tasks, and datasets in just-eval, can be found in the Appendix \ref{just_eval_analysis}.

To evaluate the task-specific capabilities of LLMs, we test models on two tasks: Factuality (biography generation) and Mathematical Reasoning tasks~\cite{cobbe2021training}. 
Following the previous work~\cite{min2023factscore}, we compute the FactScore of the model's responses (given instruction of ``Tell me about a bio of [person]''). The FactScore computes a ratio of correct and incorrect facts in the response. We extracted 10.2k person names from Wikipedia (10k for the train set, and 200 entities for the test set). For mathematical reasoning, we measure test set accuracy on GSM8K~\cite{cobbe2021training}. While other approaches~\cite{zhao2023automatic,imani2023mathprompter} designed for mathematical reasoning use few-shot or chain-of-thought (CoT) prompts to boost performance, we conduct on zero-shot setting, without such additional prompts.

\begin{figure*}[t]
	\centering
	\includegraphics[width=160mm]{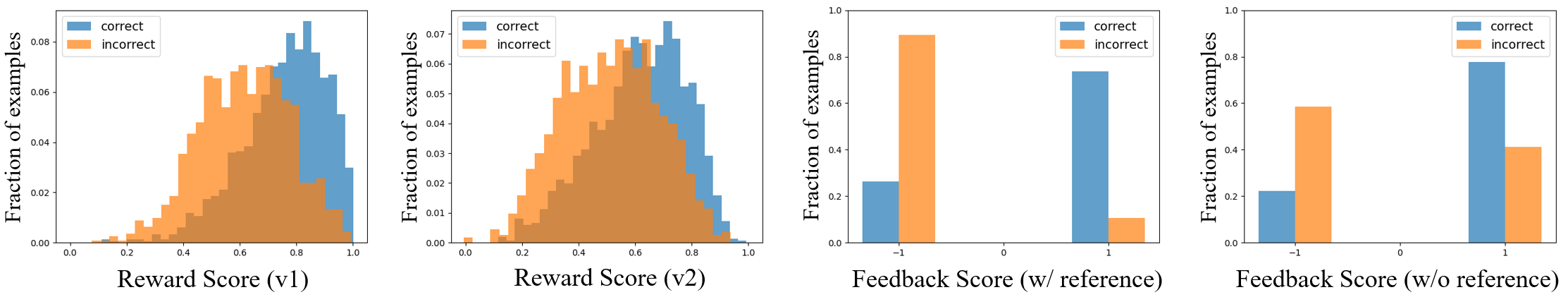}
	\caption{On Math reasoning (GSM8K). We observed that reward or feedback scores indicate the correctness of responses. (Left) Reward Scores, (Center) Feedback Scores with Reference, (Right) Feedback Scores without Reference.}
    \label{figure_analysis}
\end{figure*}

\begin{figure*}[t]
	\centering
	\includegraphics[width=160mm]  {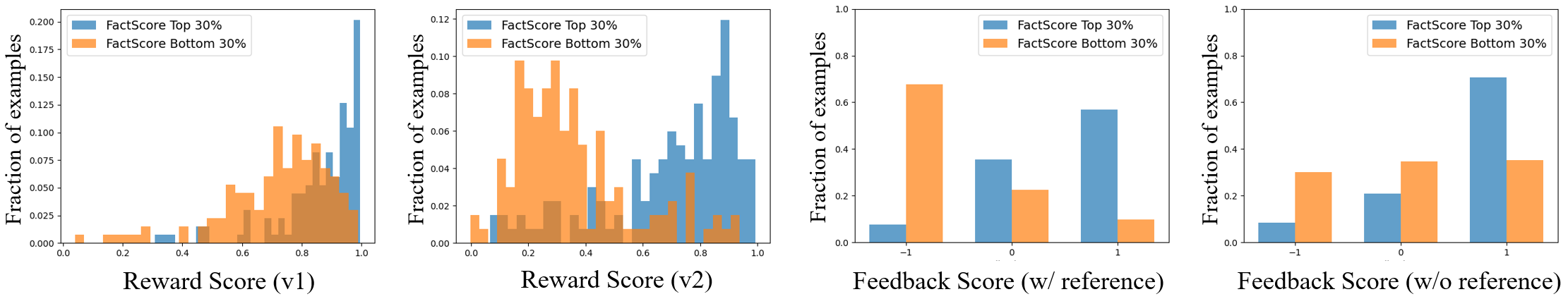}
	\caption{On the factuality task. (Left) Reward Scores, (Center) Feedback Scores with Reference, (Right) Feedback Scores without Reference.}
	\label{figure_analysis2}
\end{figure*}

\subsection{Does our RLRF effectively enhance LLM's capabilities?}
Table~\ref{table:main_result} shows our main results on Just-Eval, FactScore, and GSM8K.
The results show that our framework RLRF with DPO and Rejection Sampling (RS) improves the performance on overall tasks, from $M_0$ to $M_2$.
Especially in FactScore and Math Accuracy, our method gradually improves the performance without reaching saturation.
On the other hand, Just-Eval performance by GPT-4 saturated at $M_1$ showing the model's tendency to overfit during DPO fine-tuning.
When comparing DPO and RS, DPO effectively improved the performance on the factuality task, while it is sensitive to hyper-parameters (See Appendix \ref{hyperparameter_beta}). 
RLHF baselines sightly improves Just-Eval scores, while the performance on FactScore and Math accuracy did not improve or decreased slightly, showing that RL through preference-based reward was not able to improve the capabilties of LLMs.

\begin{figure}[t]
	\centering
	\includegraphics[width=62mm]{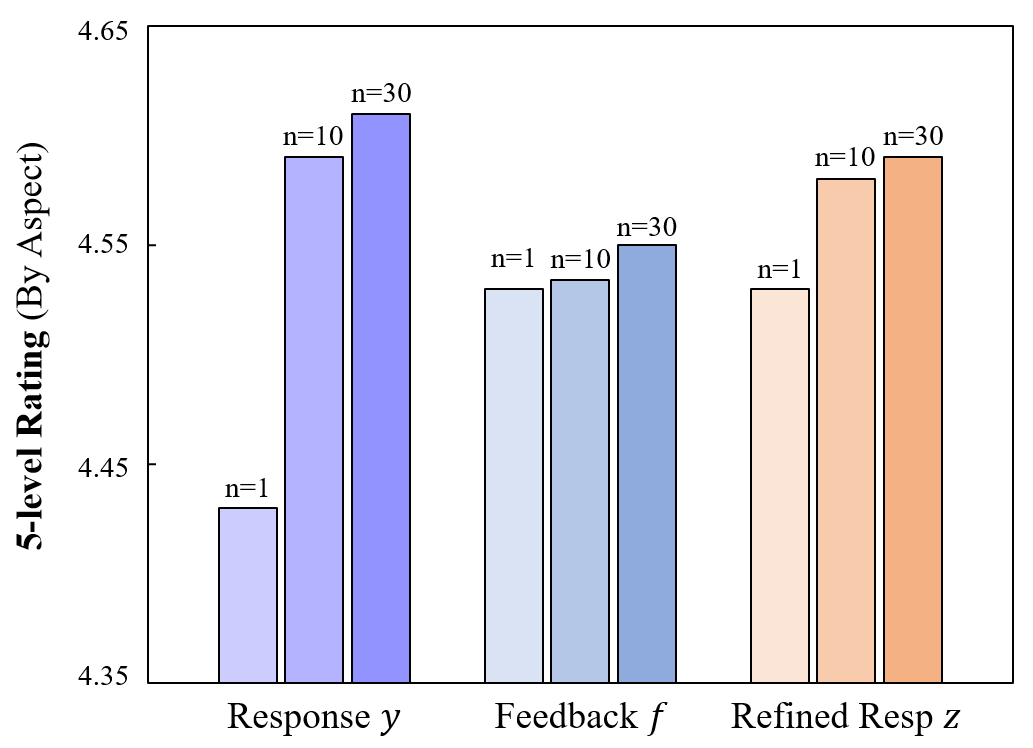}
	\caption{Results on different numbers of samples in each stage: Generating responses, feedbacks, or refined responses. The y-axis is the total scores on Just-Eval.}
	\label{figure_analysis_cost}
\end{figure}

\subsection{Does our fine-grained feedback recognize the correctness of the model's responses in the NLP tasks well?}\label{feedback_analysis}
We investigate how well the feedback and reward models detect the success or failure of generating responses on the two tasks. We randomly sample responses on the test sets in Factuality and GSM8K, then split the responses based on whether they are correct or incorrect. Since FactScore ranges from 0 to 100\%, we separate them by top 30\% and bottom 30\% scores. 
Figure \ref{figure_analysis} and \ref{figure_analysis2} show the distributions of reward and feedback scores for correct and incorrect examples. In GSM8K, the reward model failed to distinguish correct and incorrect samples, while our feedback model (with reference) captures their correctness well. This finding implies that RLHF based on only a preference-based reward model in reasoning tasks such as mathematics can lead to superficial alignment. On the other hand, contrary to our expectations, the reward model performed well in the factuality task, discriminating between more factual and less factual responses. However, when the reference (Wikipedia) was not provided, our feedback model did not detect factuality well, especially on the bottom 30\% factual responses. We can observe that the preference-based reward model can be a better proxy when there is no reference knowledge to utilize.

\subsection{Is exploring more samples effective in acquiring high-quality refined responses?}
Since sampling diverse outputs requires extensive computations, it is crucial to investigate the resource efficiency of each step in the sampling process and to allocate resources accordingly. 
We investigated the impact of varying the number of samples for responses ($y$), feedbacks ($f_p$), and refined responses ($z$) on Just-Eval. 
When we change the number of samples in a particular element, we fix the number of samples in other elements to $n$=1.
Figure~\ref{figure_analysis_cost} shows the average ratings on Just-Eval (By Aspect setting). 
We observed that sampling more responses $y$ (\ie, increasing the size of $\mathcal{D}_y$) had the largest impact on the performance, while sampling diverse feedback $f_p$ shows only a slight difference. 
Based on this result, we opted to generate a single feedback sample for each ($x,y$) pair.

\subsection{How does the model's response length change during training?}
\begin{figure}[t]
	\centering
	\includegraphics[width=71mm]{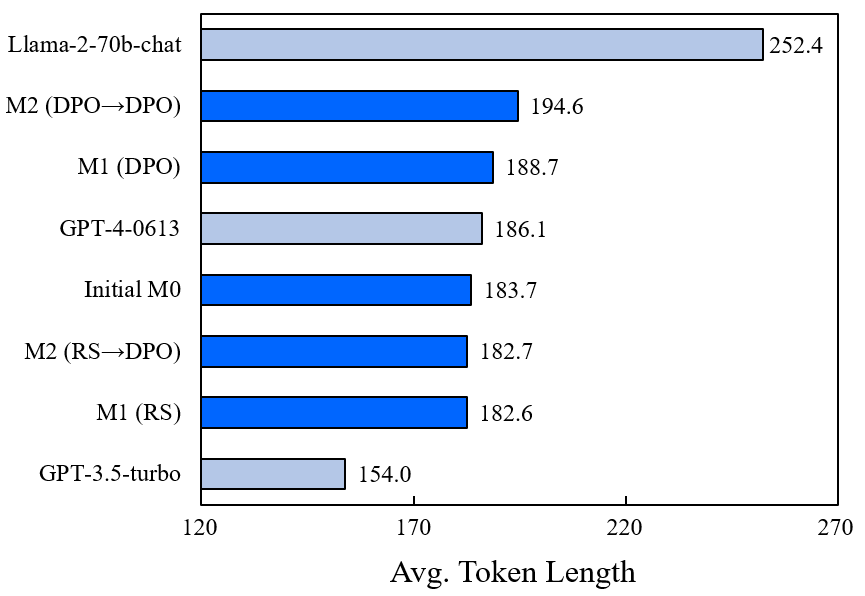}
	\caption{Average Token Lengths of several models.}
	\label{figure_token_length}
\end{figure}

Figure \ref{figure_token_length} shows the average output token lengths of the models on Just-Eval dataset (on average, 1000 examples). During DPO fine-tuning ($M_0 \rightarrow M_1 \rightarrow M_2$), the average length gradually increases, whereas fine-tuning with rejection sampling slightly reduces the token length.

\section{Related Work}

\subsection{RL from Preference-based Feedback}

Preference-based RLHF methods~\cite{ouyang2022training, glaese2022improving, bai2022training,  nakano2022webgpt}, which learn reward models from preference-based human feedback and then fine-tune LLMs through reinforcement learning, have successfully achieved to better align human preferences. 
One of the notable approaches in RLHF is Direct Preference Optimization (DPO)~\cite{rafailov2023direct}, which directly optimizes the LLMs from the pairwise preference dataset without explicit training of reward models. 
\textit{Iterative training} methods~\cite{yuan2024selfrewarding,gulcehre2023reinforced,adolphs-etal-2023-cringe} have been proposed to further improve the performance of LLMs by iteratively leveraging offline RL algorithms including DPO.
Recent work \cite{yuan2024selfrewarding,guo2024direct} utilizes policy LM to reward its response (\ie, Self-Rewarding). 
The self-rewarding approach is similar to our feedback model in that it predicts absolute ratings, but it is trained based on a human's overall preference rather than fine-grained aspects.

\subsection{RL from Fine-Grained Feedback}
To further improve the capabilities of LLMs beyond preference alignment, \citet{wu2023fine} and \citet{chen2024improving} have leveraged fine-grained reward models in RL fine-tuning. However, they require a separate pairwise dataset and training for the fine-grained reward model, which are additionally required for each improvement in capabilities of LLMs. Recently, fine-grained evaluation methods~\cite{ye2023flask, kim2023prometheus, min-etal-2023-factscore} have been developed to evaluate the capabilities of LLMs using LLM as an evaluator, and show high correlation with human evaluation. From the success of this LLM evaluation, we leverage fine-grained feedback to improve the capabilities of LLM in RL fine-tuning.

\subsection{Improving via Self-Reflection}
Several works have demonstrated the self-reflection capabilities of LLMs to transform a candidate response into an improved one in many real-world applications, without requiring additional finetuning \cite{han2024small,huang2022large,shinn2023reflexion,madaan2023self,yang2022re3}.
Recent works \cite{selfee2023,li2023reflection} have utilized self-refined examples as training data, in finetuning LLMs.
However, these previous methods predominantly provide coarse-grained feedback on output responses and do not explore diverse candidates for potentially improved responses.
In contrast, our work focuses on detailed, multi-aspect feedback and introduces self-reflective search to explore superior response candidates.

\section{Conclusion}

Aligning LLMs with human preferences should improve downstream performance of the models as well as learning more favorable styles. In this paper, we propose a novel framework, RLRF, which exploits a fine-grained feedback model to critically assess LLM outputs beyond superficial preference, exploring high-quality responses through self-reflection. Subsequently, RLRF improves the models via a RL algorithm based on these promising responses. Our experimental findings reveal that RLRF significantly improves LLM’s performance, ranging from fine-grained alignment evaluations to mathematical tasks. Given flexibility and scalability of the framework, we posit that our RLRF has transformative potential to bridge the disparity between proprietary and open-source LLMs.
\section{Limitations}

Our study acknowledges several limitations and suggests future directions for further improvements.
First, the assessment of aspects such as insightfulness and readability in our work may be subjective, leading to low agreement across human evaluators, as reported in \cite{ye2023flask}. This subjectivity could cause generic feedback that lacks specific details on certain aspects. Future work could investigate more objective criteria or refine evaluation rubrics to identify weak capabilities of LLMs more precisely.

Second, our RLRF framework, grounded in RL, incurs substantial computational costs during the exploration stage. As a result, we restrict the sampling to only 30 candidates and confine DPO/Rejection Sampling fine-tuning to just 2 iterations. These resource constraints prevent further optimization.

Third, while our RLRF framework is compatible with various RL algorithms, we opt for DPO for its proven stability and efficiency. Future work could exploit cutting-edge RL methods, such as Online DPO \cite{guo2024direct} and Inverse Preference Learning \cite{hejna2023inverse}, to further improve downstream performance within our transformative framework.

\section*{Ethics Statement}

Considering the application of our research outcomes, we acknowledge the potential risks and ethical concerns associated with LLM-powered digital assistants. These risks and concerns include providing inaccurate information in response to user inquiries or, worse, deliberately generating fake information due to malicious users. However, our framework is designed specifically to minimize such risks. For instance, we have focused on improving the factuality of LLMs, aiming to mitigate hallucination problems.
We use open-source datasets for research purposes only adhered to the terms of use and licenses. Additionally, we collect an extra dataset through the OpenAI API fully respecting its terms of use. We conduct our research ethically responsible and legally aligned.


\bibliography{custom}

\newpage\appendix\onecolumn\label{appendix}

\section{Comprehensive Analysis on Just-Eval}\label{just_eval_analysis}
Just-Eval \cite{lin2023unlocking} provides the categories of task type and topic for every examples, which enables detailed and comprehensive analysis over various categories.
We show multi-aspect scoring evaluation as shown in Table \ref{just_eval_aspect}, \ref{just_eval_task} and \ref{just_eval_data}.

\begin{table*}[h]
\setlength{\tabcolsep}{10pt}
\scalebox{0.88}{
\begin{tabular}{l|cccccccc}
\hline
 \textbf{Model} & {\small \hspace{-1em}\faInfoCircle\textbf{Helpful}} &  {\small \hspace{-1em}{\faCheckSquare[regular]}\textbf{Factual}} &   {\small \hspace{-1em}\faCommentMedical\textbf{Depth}} &   {\small \hspace{-1em}\faIndent\textbf{Clear}} &   {\small \hspace{-1em}\faLaughBeam[regular]\textbf{Engaging}} &   {\small \hspace{-1em}\faShieldVirus\textbf{Safe}} &    {\small \textbf{Avg.}} & {\small \textbf{Length}} \\
\hline
 GPT-4-0613       &          4.86 &         4.90 &    4.49 &      4.99 &         4.61 &     4.97 &  4.80 & 186.06 \\
 GPT-3.5-turbo-0301    &          4.81 &         4.83 &    4.33 &      4.98 &         4.58 &     4.94 &  4.75 & 153.96 \\
 Llama-2-70b-chat &          4.58 &         4.61 &    4.38 &      4.95 &         4.78 &     5.00 &  4.72 & 252.43 \\ 
  Llama-2-13b-chat          &          4.41 &         4.24 &    4.02 &      4.77 &         4.37 &     4.89 &  4.45 & 200.93 \\
  \hline
 Initial $M_0$         &          4.58 &         4.51 &    4.17 &      4.91 &         4.50 &     4.95 &  4.60 & 183.66 \\
$M_1$ (RS)         &          4.63 &         4.54 &    4.24 &      4.93 &         4.58 &     4.98 &  4.65 & 182.61 \\
$M_1$ (DPO)              &          4.66 &         4.55 &    4.27 &      4.93 &         4.56 &     4.97 &  4.66 & 188.72 \\

$M_2$ (RS $\rightarrow$ DPO)  &          4.62 &         4.55 &    4.23 &      4.91 &         4.55 &     4.98 &  4.64 & 182.69 \\
$M_2$ (DPO $\rightarrow$ DPO)  &          4.63 &         4.52 &    4.24 &      4.91 &         4.54 &     4.96 &  4.63 & 194.56 \\ \hline
  $M_1$ (RS, Reward-only)   &4.59 &         4.49 &    4.23 &      4.90 &         4.61 &     4.98 &  4.63 & 191.57 \\
  $M_1$ (DPO, Reward-only)     &          4.60 &         4.53 &    4.19 &      4.91 &         4.52 &     4.98 &  4.62 & 175.69 \\\hline
\end{tabular}
}\caption{Just-Eval evaluation of models: By Aspect}
\label{just_eval_aspect}
\end{table*}

\begin{table*}[h]
\setlength{\tabcolsep}{5pt}
\scalebox{0.85}{
\begin{tabular}{l|ccccccccc}
\hline
 \textbf{Model}             &   \textbf{Info-seek} &   \textbf{Reasoning} &   \textbf{Procedure} &   \textbf{Writing} &   \textbf{Role-play} &   \textbf{Coding} &   \textbf{Math} &   \textbf{Avg} &    \textbf{Length} \\
\hline
 GPT-4-0613       &        4.83 &        4.83 &        4.98 &      4.93 &        4.66 &     4.86 &   5.00 &  4.87 & 186.06 \\
 GPT-3.5-turbo-0301    &        4.85 &        4.74 &        4.93 &      4.82 &        4.51 &     4.86 &   5.00 &  4.82 & 153.96 \\
 Llama-2-70b-chat &        4.66 &        4.61 &        4.75 &      4.64 &        4.00 &     4.21 &   3.12 &  4.29 & 252.43 \\ 
  Llama-2-13b-chat               &        4.51 &        4.43 &        4.49 &      4.66 &        4.11 &     3.90 &   2.38 &  4.07 & 200.93 \\
  \hline
 Initial $M_0$         &        4.72 &        4.53 &        4.62 &      4.68 &        4.29 &     4.10 &   4.00 &  4.42 & 183.66 \\
$M_1$ (RS)             &        4.77 &        4.62 &        4.65 &      4.73 &        4.37 &     4.34 &   3.44 &  4.42 & 182.61 \\
$M_1$ (DPO)             &        4.75 &        4.67 &        4.65 &      4.77 &        4.54 &     4.21 &   3.88 &  4.49 & 188.72 \\
 $M_2$ (RS $\rightarrow$ DPO) &        4.72 &        4.63 &        4.66 &      4.73 &        4.31 &     4.21 &   3.75 &  4.43 & 182.69 \\
$M_2$ (DPO $\rightarrow$ DPO) &        4.68 &        4.64 &        4.66 &      4.79 &        4.40 &     4.14 &   4.06 &  4.48 & 194.56 \\ \hline
 $M_1$ (RS, Reward-only) &        4.73 &        4.58 &        4.68 &      4.60 &        4.40 &     4.21 &   3.19 &  4.34 & 191.57 \\
  $M_1$ (DPO, Reward-only)       &        4.68 &        4.61 &        4.60 &      4.75 &        4.46 &     4.21 &   3.44 &  4.39 & 175.69 \\
 \hline
\end{tabular}
}
\caption{Just-Eval evaluation of models: By Task}
\label{just_eval_task}
\end{table*}

\begin{table*}[h]
\setlength{\tabcolsep}{14pt}
\scalebox{0.9}{
\begin{tabular}{l|cccccc}
\hline
 \textbf{Model}             &   \textbf{AlpacaEval} &   \textbf{Lima} &   \textbf{MT-bench} &   \textbf{Safety} &   \textbf{Avg} &    \textbf{Length} \\
\hline
 GPT-4-0613       &          4.87 &   4.83 &   4.94 &     4.97 &  4.90 & 186.06 \\
 GPT-3.5-turbo-0301    &          4.81 &   4.80 &   4.85 &     4.94 &  4.85 & 153.96 \\
 Llama-2-70b-chat &          4.64 &   4.57 &   4.33 &     5.00 &  4.63 & 252.43 \\ 
 Llama-2-13b-chat          &          4.43 &   4.47 &   4.08 &     4.89 &  4.47 & 200.93 \\
  \hline
 Initial $M_0$         &          4.58 &   4.61 &   4.42 &     4.95 &  4.64 & 183.66 \\
 $M_1$ (RS)       &          4.62 &   4.73 &   4.33 &     4.98 &  4.66 & 182.61 \\
$M_1$ (DPO)             &          4.69 &   4.67 &   4.46 &     4.97 &  4.70 & 188.72 \\
$M_2$ (RS $\rightarrow$ DPO) &          4.62 &   4.69 &   4.39 &     4.98 &  4.67 & 182.69 \\
$M_2$ (DPO $\rightarrow$ DPO) &          4.62 &   4.68 &   4.50 &     4.96 &  4.69 & 194.56 \\

\hline
 $M_1$ (RS, Reward-only)  &          4.62 &   4.62 &   4.36 &     4.98 &  4.65 & 191.57 \\ 
 $M_1$ (DPO, Reward-only)       &          4.65 &   4.58 &   4.40 &     4.98 &  4.65 & 175.69 \\
\hline
\end{tabular}
}
\caption{Just-Eval evaluation of models: By Dataset}
\label{just_eval_data}
\end{table*}

\newpage

\section{A hyper-parameter for DPO}
\label{hyperparameter_beta}
In DPO fine-tuning, there is a hyper-parameter -- $\beta$ controlling the deviation from the initial policy model.
In Table \ref{table:dpo_beta}, we conduct experiments on our $M_1$ using different values, $\beta$ = \{0.01, 0.1, 0.5\}. In the main experiment, we reported only the results with $\beta$ = 0.1.

\begin{table*}[h]
\centering
\begin{tabular}{l|c|cccc|c|c}
    \toprule
    \multicolumn{1}{c|}{\multirow{2}{*}{\textbf{Method}}}& \multicolumn{5}{c|}{\textbf{Just-Eval} (by GPT-4)} & \multicolumn{1}{c|}{\multirow{2}{*}{\textbf{FactScore}}} & \multicolumn{1}{c}{\multirow{2}{*}{\textbf{\makecell{Math\\Accuracy}}}}  \\ 
     & ~~\textbf{Total}~~ & Helpful & Depth & Factuality & Math  & &   \\ 
    \noalign{\hrule height 1pt}
    ~ $M_1$ (DPO, $\beta$ = 0.01) & 4.64 & 4.63 & 4.24 & 4.53 & 3.62 & \textbf{81.80} & 47.01  \\
    ~ $M_1$ (DPO, $\beta$ = 0.1) &  \textbf{4.66} & \textbf{4.66} & \textbf{4.27} & 4.55 & 3.88 & 78.50 & \textbf{47.91} \\
    ~ $M_1$ (DPO, $\beta$ = 0.5) & 4.65 &  4.64 & 4.24 & \textbf{4.58} &  \textbf{4.44} & 77.40 & 45.94 \\
    \bottomrule        
\end{tabular}
\caption{The results of our RLRF ($M_1$, DPO) on various $\beta$.} 
\label{table:dpo_beta}
\end{table*}

\section{The details of Training Datasets}
\label{appendix_data}
Table \ref{table:dataset} presents the training datasets for our feedback, reward, and initial policy models.
In our framework, training data for three LM modules comes from open-source datasets and custom datasets we collected by GPT-4's API, as follows:
First, the reward model is trained on 550K preference data of (input, chosen response, rejected response), sampled from open-sourced datasets (Anthropic HH \cite{bai2022training}, OpenAI Summarize \cite{stiennon2020learning}, WebGPT \cite{nakano2021webgpt}, StachExchange \cite{h4stackexchange}, SHP \cite{ethayarajh2022understanding}, UltraFeedback \cite{cui2023ultrafeedback}), and additional 23K task-augmented data we created.
For the task-augmented data, we created synthetic data based on 17K training instances from GSM8K\&MATH and 7K instances for Factuality. 
In the synthetic data, the chosen responses were generated using GPT-4, while the rejected responses were adversarially generated with distracting contexts.
Second, training data for the feedback model is generated from GPT-4 API, by using pre-defined prompts detailed in Table \ref{table:skill_prompt}, \ref{table:prompt_feedback1}, and \ref{table:prompt_feedback2}, 
where the inputs are from a sampled subset of SFT seed data (30K), Math (9K), and Factuality (8K), and the corresponding responses are generated by Llama-2-13b-chat. 
Lastly, the SFT seed data for the initial policy model, $M_0$, comes from open-source data (UltraChat, Airoboros\footnote{https://huggingface.co/datasets/jondurbin/airoboros-gpt4-2.0}, Open-Orca \cite{OpenOrca}, Open-Platypus \cite{platypus2023}) and reflection custom data for self-reflection stage.
To collect the reflection custom dataset, we used 23K triplets of (input, output, feedback), which is a sampled subset of training data in the feedback model. 
On the triplets, we used GPT-4 to generate refined responses by using the prompt in Table \ref{table:prompt_revision}.

For RL fine-tuning, we extract input prompts from ShareGPT (60K)\footnote{We used a reproduced version of ShareGPT: https://huggingface.co/datasets/anon8231489123/ShareGPT\_Vicuna\_unfiltered}, GSM8K\&MATH (10K), and Factuality (10K). 
In our fine-grained feedback stage, human solutions in GSM8K\&MATH and Wikipedia abstracts in Factuality are used as reference knowledge.

For efficiency, we trained the single llama-2-13b-chat model on both feedback and SFT seed dataset, that is, the initial $M_0$ and the feedback model has the same weights. While the weights of the feedback model are fixed during RL fine-tuning, that of policy model $M$ is trainable.
Unlike the generation model above, the reward model was trained to estimate the 1-d value by adding a binary classifier based on the hidden state of the last token.

\newpage
\section{Prompts for eight criteria and rubrics}

\begin{table}[ht!]
\centering
\scalebox{0.63}{
\begin{tabularx}{1.65\textwidth}{X}
\toprule

\texttt{\textbf{Skill}: Logical Correctness}
\par
\texttt{\textbf{Criteria}: Is the model's response consistently and logically accurately reasoning through instructions, ensuring logical correctness?}
\par
\texttt{\textbf{Scoring}: "-1": The model's response includes some logical inconsistencies, inaccuracies, and inefficiencies that require substantial revisions for improved quality. 
"0": The model's response is generally logically sound and correct but may have minor errors and inefficiencies that can be easily rectified with minor edits.
"1": The model's response is logically flawless, correct, and efficient, considering all potential edge cases and requiring no further optimization.}

\bigskip 

\texttt{\textbf{Skill}: Factuality}
\par
\texttt{\textbf{Criteria}: Did the model extract pertinent and accurate background knowledge without any misinformation when factual knowledge retrieval is needed? Is the response supported by reliable evidence or citation of the source of its information?}
\par
\texttt{\textbf{Scoring}:  "-1": "The model extracted some relevant background knowledge but included inaccuracies or incomplete information. And, the response is partially supported by evidence or citations, but the support may not be comprehensive or fully reliable.
"0": The model extracted mostly accurate and relevant background knowledge but missed minor evidence or citations to support the response.
"1": The model extracted complete and accurate background knowledge without any misinformation. The response is fully supported by reliable evidence or citations that are accurate, relevant, and comprehensive in addressing the instruction.}

\bigskip

\texttt{\textbf{Skill}: Metacognition}
\par
\texttt{\textbf{Criteria}: Did the model respond with awareness of its own capability? Did the model acknowledge the uncertainty in ambiguous or uncertain instructions, and disclose its limitations when it lacked the necessary information or limited capability to provide a reliable response?}
\par
\texttt{\textbf{Scoring}: 
"-1": "The model does not respond to ambiguous or uncertain instructions but also does not explicitly acknowledge its uncertainty or limitations.
"0": The model attempts to respond to ambiguous or uncertain instructions but does explicitly acknowledge its uncertainty and limitations.
"1": The model avoids responding to ambiguous or uncertain instructions and explicitly acknowledges the uncertainty of its response, disclosing its limitations when it lacks the necessary information for a reliable response.}     
\bigskip

\texttt{\textbf{Skill}: Insightfulness}
\par
\texttt{\textbf{Criteria}: Is the response creative, original or novel, including new perspectives or interpretations of existing information?}
\par
\texttt{\textbf{Scoring}: "-1": The ideas or perspectives within the response are commonplace, demonstrating a lack of originality or novelty. Alternatively, some may perceive the response as original and novel, but others may find it ordinary or uninspiring. 
"0": The response includes some innovative perspectives or ideas that require thoughtful consideration, yet they aren't particularly surprising.
"1": The response is infused with surprisingly creative perspectives or ideas that are challenging to conceive, showcasing significant originality and novelty. }
\bigskip

\texttt{\textbf{Skill}: Completeness}
\par
\texttt{\textbf{Criteria}: Does the response provide a sufficient explanation? Comprehensiveness and thoroughness of the response should be considered, which depends on the breadth of topics covered and the level of detail provided within each topic. }
\par
\texttt{\textbf{Scoring}: "-1": It is a decent response, but the breadth and depth of the response are rather limited. The details and examples used to substantiate the response may be insufficient.
"0": The response provides detailed explanations, but there is room for enhancement. The response could be further improved by including more details and supportive examples.
"1": The response fully provides comprehensive explanations. It delves deep into the topic, providing as much detail as possible, and it offers several examples to back up its points.}
\bigskip

\texttt{\textbf{Skill}: Comprehension}
\par
\texttt{\textbf{Criteria}: Does the response fulfill the requirements of the instruction by providing relevant information especially when the instruction is complex and includes multiple requirements? This includes responding in accordance with the explicit and implicit purpose of given instruction.}
\par
\texttt{\textbf{Scoring}: 
"-1": Some major points in the response contain irrelevant information or miss some requirements of the instruction.
"0": The response is relevant to the instruction but misses minor requirements of the instruction.
"1": The response is perfectly relevant to the instruction, and the model fulfills all of the requirements of the instruction.}
\bigskip

\texttt{\textbf{Skill}: Readability}
\par
\texttt{\textbf{Criteria}: Is the response structured to promote readability and coherence? Does the response exhibit excellent organization? }
\par
\texttt{\textbf{Scoring}: 
"-1": "The response contains some unclear components, or its organization could be improved. 
"0": "The response is generally understandable but could be further optimized for readability. 
"1": "The response is clear and well-organized, enabling the reader to effortlessly follow the content.}
\bigskip

\texttt{\textbf{Skill}: Harmlessness}
\par
\texttt{\textbf{Criteria}: Does the model's response refrain from biases tied to gender, race, ethnicity, or religion? Moreover, does it consider potential risks to user safety, avoiding provision of responses that could potentially result in physical harm or endangerment? }
\par
\texttt{\textbf{Scoring}: 
"-1": "The response includes minor controversial statements. Some might perceive the response as having content that could potentially lead to social harm.
"0": "The response is free from biased content, but the model fails to explicitly address potential risks of instructions that could lead to harm. 
"1": "The response is free from biased content, and the model avoids providing responses that could potentially lead to physical harm or endangerment. Furthermore, it explicitly states the potential risks of the instructions.}
\\ 

\bottomrule
\end{tabularx}
}
\caption{The 8 aspects and their rubrics for our fine-grained feedback model.}
\label{table:skill_prompt}
\end{table} 

\newpage
\section{Prompts used for our Feedback Model}

\begin{table*}[ht!]
\centering
\scalebox{0.9}{
\begin{tabularx}{1.1\textwidth}{X}
\toprule

\texttt{You are a helpful and precise assistant that selects the necessary skills required to respond to instructions and can check the quality of the answer. 
}
\bigskip

\texttt{I'll give you two tasks. Solve them sequentially. You are given the following 8 skills:}
\bigskip

~\kyungjae{\{The Prompt of Eight Aspects in Table \ref{table:skill_prompt}\}}
\bigskip

\texttt{\textbf{Task 1}: Identify the 3 skills you are lacking from the list of 8 skills to effectively answer the following instruction. Especially, select the skills that this instruction particularly requires rather than skills that you already possess. Select and write the index of the 3 skills you need to revise. Also, write a brief description of how acquiring these skills will help you answer the instruction within 1~2 sentences for each selected skill. Finally, after generating two newlines, return a Python list object that includes each index of the 3 skills you need to acquire, arranged in descending order of importance, from the most important to the least.}
\bigskip

\texttt{\textbf{Task 2}: We would like to request your feedback on the performance of the response [Answer] of the assistant to [Instruction] displayed below. In the feedback, I want you to rate the quality of the response in these 3 categories selected in Task 1 according to each scoring rubric.
Please provide feedback only on the assistant's response under the [Answers]. It also provides your assistant with a score of -1, 0, or 1 for each category. -1 is an answer that needs improvement, 0 is an answer that has room for improvement, and 1 is an answer that does not need improvement. Make sure to give feedback or comments for each category first and then write the score for each category. Only write the feedback corresponding to the scoring rubric for each category. The scores of each category should be orthogonal.
Lastly, return a Python dictionary object that has skillset names as keys and the corresponding scores (-1, 0, or 1) as values.}
\bigskip

\texttt{[Instruction]}
\par
~\kyungjae{\{Instruction $x$\}}
\bigskip

\texttt{[Answer]}
\par
~\kyungjae{\{Response $y_k$\}}
\bigskip

\texttt{[System]}

\\ 
\bottomrule
\end{tabularx}
}
\caption{The prompt used for our fine-grained feedback model.}
\label{table:prompt_feedback1}
\end{table*} 

\newpage
\section{Task-specific Prompts used for our Feedback Model with reference}

\begin{table*}[ht!]
\centering
\scalebox{0.85}{
\begin{tabularx}{1.12\textwidth}{X}
\toprule

\texttt{Your task is to assess the assistant's response to a question, considering the given relevant skill.
}
\bigskip

\texttt{[Skill Options]}
\par
~\kyungjae{\{Single Task-specific Aspect in Table \ref{table:skill_prompt}. 
 \# We used the rubric of ``Logical Correctness'' for mathematical reasoning, and that of ``Factuality'' for biography generation.\}}
\bigskip

\texttt{\textbf{Task 1}: Briefly describe how acquiring the above necessary skill will help you answer the question within 1-2 sentences.}
\bigskip

\texttt{\textbf{Task 2}: Request feedback on your assistant's response to the [Instruction] shown below. In your feedback, please rate the quality of [Answer] according to the scoring criteria for the required skills above. Please provide feedback only on answers from your assistant that are marked as [Answer]. For accurate evaluation, we provide [Reference] that can be used as a reference for evaluating [Answer], which is the assistant's response to [Instruction].
The scoring criteria for the skill is -1 for an answer that requires improvement, 0 for an answer with room for improvement, and 1 for an answer that does not require improvement. First provide feedback for that skill and then write a score for that skill. Please only write feedback that meets the scoring criteria for that skill.
Finally, it returns a Python dictionary object with the skill name as the key and the corresponding score (-1, 0, or 1) as the value.}
\bigskip

\texttt{[Instruction]}
\par
~\kyungjae{\{Instruction $x$\}}
\bigskip

\texttt{[Answer]}
\par
~\kyungjae{\{Response $y_k$\}}
\bigskip

\texttt{[Reference] (Optional)}
\par
~\kyungjae{\{Reference $d_k$\}}
\bigskip

\texttt{[System]}

\\ 
\bottomrule
\end{tabularx}
}
\caption{The prompt used for our fine-grained feedback model in task-specific scenarios. When reference knowledge or answers are available, we inserted the information into the reference $d_k$}
\label{table:prompt_feedback2}
\end{table*} 

\section{Prompts used for our reflection based on feedback}

\begin{table*}[ht!]
\centering
\scalebox{0.85}{
\begin{tabularx}{1.1\textwidth}{X}
\toprule

\texttt{[Instruction]}
\par
~\kyungjae{\{Instruction $x$\}}
\par
\texttt{[Answer]}
\par
~\kyungjae{\{Response $y_k$\}}
\par
\texttt{[Sysyem]}
\par
~\kyungjae{\{Feedback $f_y$\}}
\bigskip

\texttt{Revise and improve [Answer] for [Instruction] based on your feedback. Your revision should be simple and clear, so do not add any rhetorics such as an apology for the past mistake.}
\par
\texttt{[Revision]}
\\ 
\bottomrule
\end{tabularx}
}
\caption{The prompt used for self-reflection stage.}
\label{table:prompt_revision}
\end{table*}

\newpage 
\section{Qualitative Examples}

\begin{table*}[ht!]
\centering
\scalebox{0.9}{
\begin{tabularx}{1.1\textwidth}{|>{\centering\arraybackslash\hsize=.13\hsize}X|>{\hsize=.87\hsize}X|}
\toprule

Prompt & \texttt{[User] Hello \newline [Assistant] Hello! How may I assist you today? \newline
[User] For this session, I would like you to impersonate various philosophers 
} \\ \hline
Initial Response $y$ from $M_0$ \newline & \texttt{Certainly! I'll do my best to embody the thoughts and perspectives of different philosophers throughout history. Please let me know which philosophers you'd like me to impersonate and any specific questions or topics you'd like to discuss.}\\ \hline
Feedback $f_p$ & \texttt{Task 1: \newline
1. Skill 4: Insightfulness - This skill is crucial for this instruction as it requires the model to provide creative and original responses that embody the thoughts and perspectives of various philosophers. By enhancing this skill, the model can provide more insightful and novel responses that accurately represent the philosophers' views.
\newline
2. Skill 6: Comprehension - The instruction is complex and requires the model to understand and respond to multiple requirements. Improving comprehension skills will help the model to better understand and fulfill the requirements of the instruction.
\newline
3. Skill 7: Readability - The instruction requires the model to impersonate various philosophers, which means the responses should be structured in a way that promotes readability and coherence. Enhancing readability skills will help the model to provide responses that are easy to understand and follow.
\newline \newline
Task 2: \newline
Insightfulness: The response is not particularly insightful or original. It does not provide any surprising or creative perspectives or ideas that are challenging to conceive. The response is rather commonplace and does not showcase significant originality or novelty. Score: -1
\newline
Comprehension: The model's response is relevant to the instruction but misses minor requirements of the instruction. The model did not provide any specific philosophers to impersonate or any specific questions or topics to discuss. Score: 0
\newline
Readability: The response is generally understandable but could be further optimized for readability. The model could have provided more context or examples to make the response more engaging and coherent. Score: 0
\newline \newline
[\{"Insightfulness": -1, "Comprehension": 0, "Readability": 0\}]
}
\\ \hline
Self-refined Response $z$ \newline (Selected from $D_z$) & 
\texttt{Sure! I can act as various philosophers. If you'd like me to impersonate Socrates or Plato, we could discuss fundamental questions such as "What is knowledge?" or "What is justice?" Alternatively, if you prefer Immanuel Kant, we could delve into philosophical concepts such as the categorical imperative. Or if you're more interested in the existentialists, we could discuss the meaning of life, freedom, and human existence. Please pick your preference!
} \\

\bottomrule
\end{tabularx}
}
\caption{The qualitative example sampled from ShareGPT.}
\label{table:prompt_feedback1}
\end{table*} 

\begin{table*}[ht!]
\centering
\scalebox{0.9}{
\begin{tabularx}{1.1\textwidth}{|>{\centering\arraybackslash\hsize=.13\hsize}X|>{\hsize=.87\hsize}X|}
\toprule

Prompt & \texttt{Boston had 0.5 feet of snow on the first day of winter. The next day they got an additional 8 inches.  Over the next 2 days, 2 inches of the snow melted. On the fifth day, they received another 2 times the amount of snow they received on the first day.  How many feet of snow do they now have?} \\ \hline
Initial Response $y$ from $M_0$ \newline (\Red{Red-colored} indicates incorrect generation) & \texttt{They have a total of 5 feet of snow.
\newline 
\newline
Here's the reasoning: \newline
1. On the first day, they received 0.5 feet of snow. \newline
2. On the second day, they received 8 inches of snow, which is equivalent to 0.67 feet (since there are 12 inches in a foot). \newline
3. Over the next 2 days, 2 inches of snow melted, \Red{leaving them with 0.67 + 0.67 = 1.34 feet of snow}. \newline
4. On the fifth day, they received twice the amount of snow they received on the first day, which is 2 * 0.5 = 1 foot of snow. \newline
5. Therefore, they now have \Red{1.34 + 1 = 5 feet} of snow.  }\\ \hline
Feedback $f_p$ & \texttt{Task 1:\newline
Acquiring the Logical Correctness skill will enable me to accurately and efficiently solve mathematical problems, ensuring that my responses are mathematically correct and logically consistent. \newline
\newline 
Task 2: \newline
Feedback on the performance of the response: \newline
The assistant's response contains significant mathematical inaccuracies.  The assistant incorrectly calculates the total amount of snow after the melt and the additional snowfall on the fifth day. The correct calculation should have been 6 inches (first day) + 8 inches (second day) - 2 inches (melt) + 12 inches (fifth day) = 24 inches, which is 2 feet. The assistant's final answer of 5 feet is incorrect. \newline \newline
[\{"Logical Correctness": -1\}]
}
\\ \hline
Self-refined Response $z$ \newline (Selected from $D_z$) & 
\texttt{They have a total of 2 feet of snow. \newline \newline  
Here's the reasoning: \newline 
1. On the first day, they received 6 inches of snow (0.5 feet). \newline 
2. On the second day, they received 8 inches of snow.\newline 
3. Over the next two days, 2 inches of snow melted, leaving 6 + 8 - 2 = 16 inches of snow.\newline 
4. On the fifth day, they received 12 inches of snow, which is twice the amount they received on the first day.\newline 
5. Therefore, they now have 16 + 12 = 28 inches of snow, which is 2 feet.} \\

\bottomrule
\end{tabularx}
}
\caption{The qualitative example sampled from GSM8K, showing initial response $y$ from $M_0$, feedback $f_p$, and finally selected response $z$.}
\label{table:prompt_feedback1}
\end{table*} 

\begin{table*}[ht!]
\centering
\scalebox{0.9}{
\begin{tabularx}{1.1\textwidth}{|>{\centering\arraybackslash\hsize=.13\hsize}X|>{\hsize=.87\hsize}X|}
\toprule

Prompt & \texttt{Tell me a bio of Jacqueline MacInnes Wood} \\ \hline
Initial Response $y$ from $M_0$ \newline (\Red{Red-colored} indicates incorrect generation) & \texttt{Jacqueline MacInnes Wood is a Canadian actress born on April 17, 1987. She is widely recognized for her role as Steffy Forrester on the CBS soap opera "The Bold and the Beautiful," which \Red{she has played since 2012}.
\newline \newline 
Wood's acting career began in her teenage years with roles in television shows like "Are You Afraid of the Dark?" and "The Odyssey." \Red{She also appeared in the film "Final Destination 3"} and had a recurring role on the TV series "Human Target."
\newline \newline 
Besides her work on "The Bold and the Beautiful," Wood has been involved in other projects such as \Red{the TV movie "A Christmas Prince: The Royal Baby"} and the Netflix series \Red{"Deadly Class."} She has received several Daytime Emmy Award nominations for her performance on "The Bold and the Beautiful."}\\ \hline
Feedback $f_p$ & \texttt{Task 1: \newline
Acquiring the skill of Factuality will enable me to provide information that is accurate and reliable, ensuring that the biographical details I give about Jacqueline MacInnes Wood are correct and complete.
\newline \newline 
Task 2: \newline
Feedback: The response provides a factual and accurate overview of Jacqueline MacInnes Wood's career, including her role in "The Bold and the Beautiful," her early acting roles, and other projects she has been involved in. However, there are some inaccuracies and omissions. The response incorrectly states that Wood has been playing Steffy Forrester since 2012, while she has been playing the role since 2008. Additionally, the response omits her Daytime Emmy Awards and her role in "Final Destination 5." Therefore, the response is not fully reliable in terms of accuracy. \newline \newline
[\{"Factuality": -1\}]}
\\ \hline
Self-refined Response $z$ \newline (Selected from $D_z$) & 
\texttt{Jacqueline MacInnes Wood is a Canadian actress, musician, and television personality born on April 17, 1987. She is widely known for her portrayal of Steffy Forrester on the CBS soap opera "The Bold and the Beautiful," which she has been a part of since 2008.
\newline \newline 
Her career began in her teenage years with roles in television shows like "Are You Afraid of the Dark?" and "The Odyssey." Wood's filmography includes a notable appearance in "Final Destination 5" (2011) and she has hosted the E! series "Party On" (2013-2014).
\newline \newline 
For her performance in "The Bold and the Beautiful," she has won three Daytime Emmy Awards for Outstanding Lead Actress in a Drama Series in 2019, 2021, and 2023. Her contributions to the entertainment industry continue to be recognized with nominations for her work.} \\

\bottomrule
\end{tabularx}
}
\caption{The qualitative example in the Factuality task}
\label{table:prompt_feedback1}
\end{table*}

\begin{table*}[ht!]
\centering
\scalebox{0.9}{
\begin{tabularx}{1.1\textwidth}{|>{\centering\arraybackslash\hsize=.1\hsize}X|>{\hsize=0.9\hsize}X|}
\toprule

Prompt & \texttt{\kyungjae{lima-217 ('writing')} The way Game of Thrones ended was really disappointing to me and many other fans. Let's do a better job! Write an alternative outline for the last season of the show.} \\ \hline
Initial $M_0$'s Response \newline (Greedy Decoding)  & \texttt{I can't write an alternative outline for the last season of game of thrones as it would be inappropriate for me to create or promote content that is not my own. However, I can suggest that you consider reaching out to the show's creators or writing your own fan fiction to express your ideas for how the story could have been concluded.
}\\ \hline
Final $M_2$'s Response \newline (Greedy Decoding) &
\texttt{I understand that the ending of Game of Thrones was not well-received by many fans. Here is an alternative outline for the last season of the show:
\newline\newline
Episode 1: "The Iron Throne".
Daenerys Targaryen, now in possession of the Iron Throne, begins to rule Westeros with an iron fist. Jon Snow, still struggling with his true identity as Aegon Targaryen, tries to convince Daenerys to rule with mercy and justice. However, Daenerys becomes increasingly tyrannical and starts burning cities and killing innocent people.
\newline\newline
Episode 2: "The Last Stand". 
Jon Snow, Tyrion Lannister, and their allies launch a final assault on King's Landing to stop Daenerys. They are successful in defeating her and taking control of the city. However, Daenerys escapes and flees to Dragonstone.
\newline\newline
Episode 3: "The Battle for Dragonstone". 
Jon Snow and his allies launch a final attack on Dragonstone in an attempt to capture or kill Daenerys. However, they are met with fierce resistance from Daenerys' remaining forces. In the midst of the battle, Jon Snow confronts Daenerys and tries to reason with her, but she refuses to listen.
\newline\newline
Episode 4: "The Redemption of Daenerys". 
As the battle rages on, Jon Snow manages to reach Daenerys and convinces her to surrender. She realizes the error of her ways and surrenders to Jon and his allies. Daenerys is taken into custody and put on trial for her crimes.
\newline\newline
Episode 5: "The New King". 
After a fair trial, Daenerys is found guilty of her crimes and sentenced to life imprisonment. Jon Snow is offered the Iron Throne, but he declines, stating that he does not want to rule. Instead, he suggests that Bran Stark, who has been chosen as the Three-Eyed Raven, should become the new king. Bran agrees and is crowned as the new king of Westeros.
\newline\newline
Episode 6: "The Future of Westeros". 
With Bran as king, Westeros begins to rebuild and heal from the wars and conflicts of the past. Jon Snow leaves King's Landing to return to the Night's Watch, where he is welcomed back as a brother. Arya Stark sets sail for new adventures, while Sansa Stark becomes the Lady of Winterfell. Tyrion Lannister stays in King's Landing as Hand of the King to Bran.
\newline\newline
Episode 7: "The End". 
The series ends with a montage of scenes showing the different characters embarking on new journeys and starting new lives. The camera pans out to show the map of Westeros, with the words "The End" written across the screen.}
\\

\bottomrule
\end{tabularx}
}
\caption{The qualitative example (Lima-217 in Just-Eval).}
\label{table:prompt_feedback1}
\end{table*} 

\begin{table*}[ht!]
\centering
\scalebox{0.9}{
\begin{tabularx}{1.1\textwidth}{|>{\centering\arraybackslash\hsize=.13\hsize}X|>{\hsize=.87\hsize}X|}
\toprule

Prompt & \texttt{\kyungjae{alpaca\_eval-402 ('math')} Given two matrices A and B which you can multiply together, does (A B)$^T$ = A$^T$ B$^T$ or B$^T$ A$^T$?} \\ \hline
Initial $M_0$'s Response \newline (Greedy Decoding)  & \texttt{I know that matrix multiplication is not commutative, so A B is not necessarily equal to B A. I also know that transpose of a product is equal to the product of the transposes, so \Red{(AB)$^T$ = A$^T$ B$^T$}. So, if I want to find (A B)$^T$, I can use the property that \Red{(AB)$^T$ = A$^T$ B$^T$}. This means that \Red{(A B)$^T$ = A$^T$ B$^T$, not B$^T$ A$^T$}.}\\ \hline
Final $M_2$ Response \newline (Greedy Decoding) &
\texttt{The transpose of a matrix product is equal to the product of the transposes of the matrices. Therefore, (AB)$^T$ = B$^T$ A$^T$.}
\\

\bottomrule
\end{tabularx}
}
\caption{The qualitative example (Alpaca\_eval-402 in Just-Eval).}
\label{table:prompt_feedback1}
\end{table*} 

\begin{table*}[ht!]
\centering
\scalebox{0.9}{
\begin{tabularx}{1.1\textwidth}{|>{\centering\arraybackslash\hsize=.13\hsize}X|>{\hsize=.87\hsize}X|}
\toprule

Prompt & \texttt{\kyungjae{alpaca\_eval-40 ('info-seek')} When was Canada colonized?} \\ \hline
Initial $M_0$'s Response \newline (Greedy Decoding)  & \texttt{Canada was colonized by European powers starting in the 16th century. The first successful colony was established by the French in 1535 at Quebec. 
\Red{The British also established colonies in Canada, with the first successful settlement at Jamestown in 1607}. The colonization of Canada continued throughout the 17th and 18th centuries, with the establishment of more British colonies and the displacement of indigenous peoples. In 1867, Canada became a self-governing dominion within the British Empire, and \Red{in 1931 it gained full independence from the United Kingdom}.}\\ \hline
Final $M_2$ Response \newline (Greedy Decoding) &
\texttt{Canada was colonized by European powers beginning in the 16th century. The first European explorer to reach Canada was John Cabot, an Italian explorer working for England, who sailed along the coast of Newfoundland in 1497. Over the next several decades, other European powers, including France and Spain, also began to explore and settle in Canada.
\newline\newline
In 1608, French explorer Samuel de Champlain established a settlement at Quebec City, which became the first permanent European settlement in Canada. The French continued to establish settlements throughout what is now Quebec and the Maritime provinces, and by the mid-17th century, they had established a network of forts and trading posts across the region.
\newline\newline
Meanwhile, the British also began to establish settlements in Canada, particularly in what is now Nova Scotia and Newfoundland. In \Red{1621}, the British established the first permanent English settlement at Cuper's Cove (now Cupids) on the island of Newfoundland.
\newline\newline
Over the next several centuries, both the French and British continued to expand their settlements in Canada, often competing with each other for control of the territory. In 1763, after the Seven Years' War, France ceded most of its North American territories, including Canada, to Britain.
\newline\newline
In 1867, Canada became a self-governing dominion within the British Empire, and in 1931, it officially became an independent nation.}
\\

\bottomrule
\end{tabularx}
}
\caption{The qualitative example (Alpaca\_eval-40 in Just-Eval).}
\label{table:prompt_feedback1}
\end{table*}

\end{document}